\documentclass[]{article}
\usepackage[]{geometry}
\usepackage{amta2020}
\usepackage{times}
\usepackage{url}
\usepackage{latexsym}
\usepackage{natbib}
\usepackage{layout}
\usepackage{amsmath}
\usepackage{amssymb,amsfonts}
\usepackage{epsfig}

\usepackage{dsfont}
\usepackage{mathtools}
\usepackage{xcolor}
\usepackage{multirow}
\usepackage[draft,marginpar]{todo}
\usepackage{xspace}

\newcommand{\model}[1]{\texttt{#1}}

\newcommand{\jsrc}{\ensuremath{f_1^J}} 
\newcommand{\itrg}{\ensuremath{e_1^I}} 
\newcommand{\ztrg}{\ensuremath{e_0^I}} 
\newcommand{\jali}{\ensuremath{a_1^J}} 
\newcommand{\ibli}{\ensuremath{b_1^J}} 

\newcommand{\expectP}[2]{\ensuremath{\mathbb{E}_{#1}(#2)}}

\newcommand{\fyDone}[1]{\done[FY]\Todo[FY:]{\textcolor{orange}{#1}}}
\newcommand{\fyFuture}[1]{\done[FY]\Todo[FY:]{\textcolor{red}{#1}}}

\newcommand{\akDone}[1]{\done[AK:]{\textcolor{blue}{#1}}}

\begin{document}

\title{\bf Generative latent neural models for automatic word alignment}

 \author{\name{\bf Anh Khoa Ngo Ho} \hfill  \addr{anh-khoa.ngo-ho@limsi.fr}\\
         \name{\bf Fran\c{c}ois Yvon} \hfill \addr{francois.yvon@limsi.fr}\\
         \addr{Universit\'e Paris-Saclay, CNRS, LIMSI \\ B\^{a}t. 508, rue John von Neumann, Campus Universitaire, F-91405 Orsay}
}


\maketitle
\pagestyle{empty}

\begin{abstract}
\fyDone{Write abstract}
Word alignments identify translational correspondences between words in a parallel sentence pair and are used, for instance, to learn bilingual dictionaries, to train statistical machine translation systems or to perform quality estimation. Variational autoencoders have been recently used in various of natural language processing to learn in an unsupervised way latent representations that are useful for language generation tasks. In this paper, we study these models for the task of word alignment and propose and assess several evolutions of a vanilla variational autoencoders. We demonstrate that these techniques can yield competitive results as compared to Giza++ and to a strong neural network alignment system for two language pairs.
\end{abstract}

\section{Introduction
\label{ssec:introduction}}
Word alignment is one of the basic tasks in multilingual Natural Language Processing (NLP) and is used to learn bilingual dictionaries, to train statistical machine translation (SMT) systems \citep{Koehn10statistical}, to filter out noise from translation memories \citep{Pham18fixing} or in quality estimation applications \citep{Specia18quality}. Word alignments can also be viewed as a form of possible explanation of the often opaque behavior of a Neural Machine Translation \citep{Stahlberg18operation}. Word alignment aims to identify translational equivalences at the level of individual lexical units \citep{Och03asystematic,Tiedemann11bitext} in parallel sentences. 

Successful alignment models either rely on bilingual association measures parameterizing a combinatorial problem (eg.\ an optimal matching in a bipartite graph); or on probabilistic models, as represented by the IBM Models of \citet{Brown1993Mathematics} and the HMM model of \citet{Vogel96HMM}. All these models use unsupervised learning to estimate the likelihood of alignment links at the word level from large collections of parallel sentences.

Such approaches are typically challenged by low-frequency words, whose co-occurrences are poorly estimated; they also fail to take into account context information in alignment; finally, they make assumptions that are overly simplistic (eg.\ that all alignments are one-to-many or many-to-one), especially when the languages under focus belong to different linguistic families. Even though their overall performance seem fair for related languages (eg.\ French-English), there is still much room for improving.  Indeed, the error rate of automatic alignments tools such as \model{Giza++} \citep{Och03asystematic} or \model{Fastalign} \citep{Dyer13asimple},  even for high resource languages, is still well above 15-20\%; and the situation is much worse in low-resource settings \citep{Martin05word,Xiang10diversify,Mccoy18phonologically}. 

As for most NLP applications \citep{Collobert11Natural}, and notably for machine translation \citep{Cho14ontheproperties,Bahdanau15neural}, neural-based approaches offer new opportunities to reconsider some of these issues.
Following up on the work of eg.\ \citep{Yang2013Word,Alkhouli16alignmentbased,Wang2018Neural}, we study ways to take advantage of the flexibility of neural networks to design effective variants of generative word alignment models.

Our main source of inspiration is the model of \cite{Rios18deepgenerative}, who consider variational autoencoders \citep{Kingma14autoencoding, Rezende14stochastic} to approach the unsupervised estimation of neural alignment models. We revisit here this model, trying to analyze the reasons for its unsatisfactory performance and we extend it in several ways, taking advantage of its fully generative nature. We first generalize the approach, initially devised for IBM model 1, to the HMM model; we then explore ways to effectively enforce symmetry constraints \citep{Liang06alignment}; we finally study how these models could benefit from monolingual data. Our experiments with the English-Romanian and English-French\fyDone{Check this} language pairs show that our best model with symmetry constraints is on par with a conventional neural HMM model; they also highlight the remaining deficiencies of these approaches and suggest directions for further developments.     

\section{Neural word alignment variational models}
The standard approach to probabilistic alignment \citep{Och03asystematic} is to consider \emph{asymmetric} models associating each word in a source sentence $\jsrc = f_1\dots{} f_J$ of $J$ words with exactly one word from the target sentence $\ztrg = e_0 \dots{} e_I$ of $I+1$ words.\footnote{As is custom, target sentences are completed with a ''null'' symbol, conventionally at index $0$.} This association is governed by unobserved alignment variables $\jali = a_1 ... a_J$, yielding the following model:
\begin{align}
  p(\jsrc,\jali|\ztrg) &= \prod_{j}^J p(a_j|a_1^{j-1},f_1^{j-1},\ztrg) p(f_j|a_1^{j},f_1^{j-1},\ztrg)
\end{align}
Two versions of this model are considered here: in the IBM model~1 \citep{Brown1993Mathematics}, the alignment model $p(a_j|a_1^{j-1},f_1^{j-1},\ztrg)$ is uniform; in the HMM model of \cite{Vogel96HMM}, Markovian dependencies between alignment variables are assumed and $a_j$ is independant from all the preceding alignment variables given $a_{j-1}$. In both models, $f_j$ is conditionally independent to any other variable given $a_j$ and $\itrg$. Under these assumptions, both parameter estimation and optimal alignment can be performed efficiently with dynamic programming algorithms. In this approach, $\itrg$ is not modeled. 

\subsection{A fully generative model}
We now present the fully generative approach introduced by \citet{Rios18deepgenerative}. In this model, the association between a source word $f_j$ and a target word $e_i$ is mediated by a shared latent variable $y_i$, assumed to represent the joint underlying semantics of mutual translations. In this model, the target sequence $\itrg$ is also modeled, yielding the following generative story:\footnote{We omit the initial step, consisting in sampling the lengths $I$ and $J$ and the dependencies wrt. these variables.}
\begin{enumerate}
\item Generate a sequence $y_0^I$ of d-dimensional random embeddings by sampling independently from some prior distribution e.g. Gaussian 
\item Generate $\itrg$ conditioned on the latent variable sequence $y_1^I$
\item Generate $\jali = a_1 ... a_J$ denoting the alignment from \jsrc{} to  $y_0^I$ 
\item Generate $\jsrc$ conditioned on $y_0^I$ and $\jali$
\end{enumerate}
This yields the following decomposition of the joint distribution of \jsrc{} and \itrg{}, where we marginalize over latent variables $y_0^I$ and $\jali$: 
\begin{align}
p(\jsrc, \itrg{}) 
&= \!\! \int_{y_0^I} \!\!p(y_0^I) p_{\theta}(\itrg| y_1^I) \big( \sum_{a_1^J} p_{\theta}(\jali) p_{\theta}(\jsrc| y_0^I,\jali) \big) dy_0^I
\label{eq:marginal_likelihood}
\end{align}
\fyDone{Consistency in notations P, $p_\theta$, p}
Directly maximizing the log-likelihood to estimate the parameters is in general intractable, especially when neural networks are used to model the generation of $\jsrc$ and $\itrg$. The standard approach in neural generative models \citep{Kingma14autoencoding} is to introduce a variational distribution $q_\phi$ for the latent variables and to optimize the so-called evidence lower-bound (ELBO). Following \citep{Rios18deepgenerative} we consider tractable alignment models and use the variational distribution only for modeling $y_0^I$ conditioned on $\itrg$. This yields the following objective:
\begin{align}
J(\theta, \phi)= &- \expectP{q_{\phi}(y_1^I)}{\log\,p_{\theta}(\itrg| y_1^I)} - \expectP{q_{\phi}(y_0^I)}{[ \log \sum_{a_1^J}p_{\theta}(\jali) p_{\theta}(\jsrc| y_0^I, \jali)} \nonumber\\[-4pt]
&\quad\quad + \operatorname{KL}[q_{\phi}(y_0^I|\itrg)||p(y_0^I)]
\label{eq:objective_function_main}
\end{align}
where $\expectP{p}{f}$ denotes the expectation of $f$ with respect to $p$, and $\operatorname{KL}$ is the Kullback-Leibler divergence.
Objective~\eqref{eq:objective_function_main} is a sum of three terms that are referred respectively as the \emph{reconstruction cost}, the \emph{alignment cost} and \emph{KL divergence cost}. The last term can be computed analytically when the prior and the variational distributions are Gaussian and we thus assume the following parameterization $q_\phi(y_1^I | e_1^I) = \prod_{i} N(y_i | u_i, s_i)$, where the mean $u_i$ and the diagonal covariance matrix $\operatorname{diag}(s_i)$ are deterministic functions of $\itrg$. As is custom, the expectations in equation~\eqref{eq:objective_function_main}\fyDone{use eqref} are approximated by sampling values of $y_i$ as $y_i = u_i + s_i \cdot \epsilon_i$, where $\epsilon_i$ is drawn from a white Gaussian noise.\fyDone{Check notation} The reparameterization trick removes the sampling step from the generation path, and makes the whole objective differentiable \citep{Kingma14autoencoding}. 

\subsection{Introducing Markovian dependencies \label{ssec: HMM+VAE}}

The experiments in \citep{Rios18deepgenerative} only consider basic assumptions regarding the alignment model $p_\theta(\jali)$,\fyDone{notations} corresponding to IBM model 1. Our first variation of this model considers a richer transition model assuming Markovian dependencies, for which the exact marginalization of alignment variables implied by equation~\eqref{eq:objective_function_main} remains tractable with the forward algorithm. The alignment cost is the expectation of the source given the latent variables:
\fyDone{use $\log$ througout}\fyDone{Check equation : no outer sum}
\begin{align}
  \expectP{q_{\phi}(y_0^J)}{[ \log \sum_{a_1^J} \prod^J_{j=1} p_\theta(f_j| y_{a_j}) p_\theta(a_j| a_{j-1})]}
\label{eq:hmm_alignment}
\end{align}
As is usual with HMM variants of alignment models, we parameterize the transition distribution $p_\theta(a_j| a_{j-1})$ on the distance (jump) between the values of $a_j$ and $a_{j-1}$ \citep{Och03asystematic}. This model is referred to below as \model{HMM+VAE}.\fyFuture{Check representations}

\subsection{Towards symmetric models: a parameter sharing approach \label{ssec:VAE+SP}}

\fyDone{titles}A first benefit of having a fully generative model (in both alignment directions), which jointly models $\jsrc$ and $\itrg$, is that it becomes easy to encourage these models to share information and to improve their joint performance.
Our alignment models involves two decoders, one for the source and one for the target (in each direction). These components are used to compute a distribution over vocabulary words given a d-dimensional variable, and are conceptually similar.

Our first step is thus to simultaneously train the alignment models in both directions, making sure that they use the same decoder respectively for $\jsrc$ and $\itrg$. This means that the same network computes $p_{\theta}(\itrg|y_1^I)$ (when $\itrg$ is in the target) and $p_{\theta}(\itrg|y_0^J,\jali)$ when $\itrg$ is the source.\fyDone{Fix notations}
There is only one encoder computing the variational parameters in each direction, and these remain distinct in this approach.
Our joint objective function now comprises six terms including two reconstruction costs, two alignment costs and two KL divergence costs. From this, we see that a first benefit of this method is computational as it greatly reduces the number of parameters to train. We also expect that it will yield two additional benefits: (a) to help improve the alignment model, which is more difficult to train for lack of observing the ``right'' alignment variables; in comparison the reconstruction of the target sentence is almost obvious, as each $e_i$ is generated from the right $y_i$; (b) to make the alignments more symmetrical, thereby facilitating their interpretation and their recombination. This model is denoted \model{+VAE+SP} below.\fyFuture{Look at the embeddings.}

\subsection{Enforcing agreement in alignment \label{ssec:+VAE+SP+AC}}
The idea of training two asymmetrical models opens new ways to control the level of agreement between alignments, an idea already considered eg.\ in \citep{Liang06alignment,Graca10learning}. Following the former approach, we implement this idea by adding an extra cost that rewards agreement between asymmetric alignments. For non null alignment links, this cost is based on the alignment posterior distributions and is defined as:
\begin{equation}
\sum_{i>0,j>0} |p(a_j = i| \jsrc, \itrg) - p(b_i = j| \jsrc, \itrg)|,
\end{equation}
where \ibli{} is the alignment variables introduced when $\itrg$ is the source of the alignment, and $\jsrc$ is the target. Both for the IBM-1 and for the HMM variants, these posterior distributions can be computed effectively, in the latter case using the forward-backward algorithm.

In the case of the null links, the agreement term should reward configurations where one source word is aligned with the null symbol in one direction, and is not aligned to any target word in the other direction. This yields the following additional term (for the canonical source to target direction, the reverse term is analogous):
\begin{equation}
  \sum_{j=1}^J |1 - p(a_j = 0|\jsrc, \itrg) - \sum_{i=1}^I p(b_i = j|\jsrc, \itrg)|
\end{equation}
For this model (\model{+VAE+SP+AC}), the objective function comprises nine terms, each with its own dynamics, which makes optimization more difficult due to the heterogeneity between costs.\fyDone{Je ne comprends pas}

\subsection{Training with monolingual data \label{ssec:+Mono}}
\akDone{}
Leaving the alignment module aside, the model can be used as a simple autoencoder which can be (pre)trained monolingually. We use supplementary monolingual sentences $\dot{e}_1^M$ that just go through the encoding-decoding process, and add an extra monolingual reconstruction term $J_{\operatorname{mono}}$ in the objective \eqref{eq:objective_function_main}:
\begin{align}
J_{\operatorname{mono}}(\theta, \phi)= - \expectP{q_{\phi}(\dot{y}_1^M)}{\log\,p_{\theta}(\dot{e}_1^M| \dot{y}_1^M)} + \operatorname{KL}[q_{\phi}(\dot{y}_1^M|\dot{e}_1^M)||p(\dot{y}_1^M)]
\end{align}
where $\dot{y}_1^M$ is the latent variable associated to $\dot{e}_1^M$.
Alternatively, we consider training the alignment model monolingually. We implement this idea by adding a random noise to the target sentence, to make it more similar to a source sentence and amenable to alignment. In this case, the extra reconstruction term is:
\begin{align}
J_{\operatorname{mono}}(\theta, \phi)= - \expectP{q_{\phi}(\ddot{y}_0^N)}{[ \log \sum_{\ddot{a}_1^M}p_{\theta}(\ddot{a}_1^M) p_{\theta}(\dot{e}_1^M| \ddot{y}_0^N, \ddot{a}_1^M)} + \operatorname{KL}[q_{\phi}(\ddot{y}_0^N|\ddot{e}_1^N)||p(\ddot{y}_0^N)]
\end{align}
where $\ddot{e}_1^N$ is a noisy version of $\dot{e}_1^M$, $\ddot{y}_1^N$ is the latent variable for $\ddot{e}_1^N$, and $\ddot{a}_1^M$ denotes the alignment variables betwen $\dot{e}_1^M$ and  $\ddot{y}_0^N$. In our experiments, we only use IBM Model 1 as our alignment model.

\section{Experiments} \label{sec:experiments}
\subsection{Datasets \label{ssec:datasets} }
Our experiments use two standard benchmarks from the 2003 word alignment challenge \citep{Mihalcea2003Evaluation}, respectively for aligning English with French and Romanian. We consider two different settings: for French, we use a large training corpus of parallel sentences from the Europarl corpus \cite{Koehn2005Europarl}. In the case of Romanian, we use the SETIMES corpus used in WMT'16 evaluation,\footnote{http://statmt.org/wmt16} which correspond to a more challenging scenario where the training data is limited in size.\fyFuture{Real tests with the 2005 data}
\fyDone{How about monolingual data?} Additional experiments with monolingual data use the Romanian data from News Crawl 2019 ($\sim$ 6M sentences)\footnote{See http://statmt.org/wmt19}. Basic statistics for these corpora are in Table~\ref{tab:corpus_size}.

\begin{table}[h!]
  \begin{center}
    \begin{tabular}{ | c | c | c | c | c| c |}
      \hline
      Corpus & \# sent. in train & \# sent. in test &  \multicolumn {2}{c|}{\# tokens in test} & \# non-null links \\
            & & & Eng. & For. & \\
      \hline \hline
      En-Fr & $\sim$1.9M & 447  & 7~020 & 7~761 & 17~438\\
      \hline
      En-Ro & $\sim$260K & 246 & 5~455 & 5~315 & 5~988\\ 	 
      \hline
    \end{tabular}
    \caption{Basic statistics for the data}
    \label{tab:corpus_size}
  \end{center}
\end{table}

These corpora are preprocessed, lowercased and tokenized with standard tools from the Moses toolkit.\footnote{https://github.com/moses-smt/mosesdecoder}\fyDone{tokenization ? truecasing ?} Following notably \citep{Garg19jointlylearning}, we perform the alignment between subword units generated by Byte-Pair-Encoding \citep{Sennrich2015Neural}, implemented with the SentencePiece model \citep{Kudo2018Sentencepiece} and computed independently\footnote{We differ there from \cite{Garg19jointlylearning} who use a joint BPE vocabulary.} in each language with 32K merge operations. This makes the training less computationally demanding and greatly mitigates the rare-word problem, which is a major weakness of historical count-based model. Our results and analyses are however based on word-level alignments. Subword-level alignments are converted into word-level alignments as follows: a link between a source and a target word exists if there is at least one link alignment between their subwords.

\subsection{Implementation \label{ssec:implementation}}
Our models are close in structure to the model proposed by \citet{Rios18deepgenerative}, and are made of three main components: an encoder to generate the latent variables $y_0^I$ from $\itrg$, and two decoders to respectively reconstruct $\itrg$ and $\jsrc$, with the help of the alignment model.

The encoder is composed of a token embedding layer (128~units), two LSTM layers (each comprising 64 units), and dense output layers to independently generate the mean vectors ($u_1 \dots u_I$) vectors and the diagonal of the covariance matrices ($s_1\dots{}s_I$). The latent variable $y_1^I$ has 64 units.\footnote{In our BPE baseline experiments with En:Ro, we found that 64 hidden units were sufficient to obtain the best AER score after 10 iterations. As for the other meta-parameters, we decided to stick with these baseline values.}\fyDone{If each LSTM layer has 64 cells, we should have 128 at the output of the biLSTM ? There is a layer to $W_h$ to transform the concatenate into the size required} Our encoder is formally defined as:
\begin{align*}
\begin{split}
&\overrightarrow{h_i} = RNN(\overrightarrow{h_{i-1}}, E(e_i))\\
&h_i = W_{h} \operatorname{concat}(\overrightarrow{h_i}, \overleftarrow{h_i} ) \\
\end{split}
\begin{split}
& s_i =  \operatorname{softplus}(W_{s} h_i + b_{s} )\\
& u_i = W_{u} h_i + b_{u}  \\
& y_i = u_i + s_i \cdot \epsilon_i
 \end{split}
\end{align*}
\fyDone{Fix cdot}where $E(e_i) \in \mathbb{R}^{128}$ is the embedding of word $e_i$,
$\epsilon$ is a noise variable $\epsilon \sim N(0,1)$ and $\operatorname{softplus} = \log(1+\exp(x))$ is an activation function returning a value positive. The vector $y_0$ is independently generated from a pseudo-sentence made of one dummy token; it is identical for all target sentences. Note that the decoder model does not try to reconstruct this token. 
\fyDone{Expliquer $y_0$}
The reconstruction decoder is given by:
\begin{align*}
p_\theta(e_i|y_i) = [\operatorname{softmax}(W_{v} y_i + b_{v})]_{e_i},
\end{align*}
and the alignment model with emission and transition components is:
\begin{align*}
&p_\theta(f_j|e_{a_j}) = [\operatorname{softmax}(W_{v} y_{a_j})]_{f_j} \\
&p_\theta(a_j - a_{j-1}) = [\operatorname{softmax}(W_{\Delta} y_{a_{j-1}})]_{a_j - a_{j-1}}
\end{align*}
where $W_{v} \in \mathbb{R}^{64 \times V}, b_{v} \in \mathbb{R}^{V}$, with $V$ the target vocabulary size. $W_{\Delta} \in \mathbb{R}^{64 \times 301}$ with jump values in the interval  $[-150, +150]$.\fyDone{Ecrire le modele d'alignement correct}
\akDone{}

For experiments with monolingual data, our noise model follows the technique in \citep{Lample2017Unsupervised}. We randomly delete input words with probability $p_{wd} = 0.1$. We then slightly shuffle the sentence, where the difference between the position before and after shuffling each word is smaller than $4$.

In all cases, our optimizer is Adam \citep{Kingma2014Adam} with an initial learning rate of 0.001; the batch size is set to 100 sentences. We use all training sentences of length lower than 50. All parameters of the \model{Giza++} and \model{Fastalign} baselines are set to their default values. \model{IBM-1+NN} and \model{HMM+NN} correspond to basic neuralizations of the IBM models as in \citep{Rios18deepgenerative,Ngoho19neural} for both word-level and BPE-level. These models are trained by maximizing the likelihood with the expectation-maximization algorithm. We train all models for 10~iterations. Results with symmetric alignments use the grow-diag-final (GDF)\fyDone{or just GDF ?} heuristic proposed in \citep{Koehn05Edinburgh}. 

\subsection{Evaluation protocol \label{ssec:evaluation_protocol}}
We use the alignment error rate (AER) \citep{Och2003Minimum}, accuracy,\fyDone{Is this useful ?} F-score, precision and recall
as measures of performance. AER is based on a comparison of predicted alignment links (A) with a human reference including sure (S) and possible (P) links, and is defined as an average of the recall and precision taking into account the sets $P$ and $S$. AER is defined as:
 \begin{align*}
 AER & = 1 - \dfrac{|A \cap S| + |A \cap P|}{|A| + |S|}
 \end{align*}
where $A$ is the set of predicted alignments. Note that the Romanian-English reference data only contains sure links; in this case AER and F-measure are deterministically related.


\subsection{Results} \label{ssec:results}
The top part of Table~\ref{tab:AER_IBM-1} reports the AER score of the IBM-1 baselines: the count-based model (\model{IBM-1 Giza++}) and the two neural variants, operating at the word (\model{IBM1+NN}) and subword (\model{IBM-1+BPE}) levels. We also report the performance of three VAE variants (\model{IBM1+VAE+BPE}, \model{IBM1+VAE+BPE+SP}, \model{IBM-1+VAE+BPE+SP+AC}). A first observation is neural baselines are better than Giza++, and that using BPE units brings an additional gain.

The basic model (\model{IBM-1+VAE}) falls short to match these results and proves way worse than the two neural version of the IBM-1 model. These results are in line with the findings of \cite{Rios18deepgenerative}, who report similar difference in performance. Sharing the parameters between directions greatly improves this baseline with a reduction in AER of about 8 points (En-Fr) and 6 points (En-Ro) for both directions, as well as for symmetrization. The reconstruction model, which is well trained in one direction, helps to improve the emission model in the reverse direction. We observe that the gain is more significant when the morphologically rich language is on the target side: this is were the emission model is the weakest and benefits most from parameter sharing. Adding an extra agreement cost fails to produce markedly better alignments for Fr-En; we however observe a gain of about 2 AER points for the symmetricized alignments in En-Ro. Overall, our best VAE model outperform the neural baseline in the large training condition (English-French); we do not see this for the other language pair, where the performance remains much below the neural baseline.

\begin{table}[h!]
\center
\begin{tabular}{ l  |c c c | c c c }
 Model & \multicolumn{3}{ c}{English-French}
 & \multicolumn{3}{|c }{English-Romanian} \\ 
\cline{2-7} 
 \model{IBM-1} & En-Fr & Fr-En & GDF & En-Ro & Ro-En & GDF \\ 
\hline 
 \model{Giza++} & 40.0 & 33.9 & 25.1 & 56.0 & 53.5 & 51.1\\
\hline 
 \model{IBM1+NN}     & 27.9 & 27.2 & 17.8 & 46.3 & 44.9 & 38.3 \\ 
 \model{IBM1+NN+BPE}    & 25.7 & 24.0  & \textbf{14.6} & \textbf{43.4} & \textbf{40.4} & \textbf{34.4} \\ \hline 
 \model{IBM1+VAE+BPE}    & 33.4 & 34.3 & 24.9 & 56.3 & 55.6 & 51.3 \\ 
 \model{+SP}      & \textbf{22.1}  & 23.8 & 16.8  & 49.3 & 51.4 & 45.2 \\ 
 \model{+AC}      & 22.8 & \textbf{23.6} & 17.8  & 49.1 & 49.2 & 43.3 \\ 
\hline 
\end{tabular} 
\caption{AER scores for \model{IBM-1} models. The best result in each column is in boldface.}
\label{tab:AER_IBM-1}
\end{table}

The effect of adding a transition component in these models is less clear, as shown in Table~\ref{tab:AER_HMM}, where we report the performance of HMM-based variants. Both symmetrization strategies prove again very effective to improve the basic VAE model, and our best system (\model{+AC}) achieves AER scores that are close, yet slightly inferior, to the \model{HMM+NN+BPE} baseline. One possible issue that we do not fully solve via symmetrization is related to the null word, which, as explained above, is not part of the reconstruction model, and which does not improve with joint learning.


\begin{table}[h!]
\center
\begin{tabular}{ l  |c c c | c c c }
 Model & \multicolumn{3}{ c}{English-French} 
 & \multicolumn{3}{|c }{English-Romanian} \\ 
\cline{2-7} 
 \model{HMM}  & En-Fr & Fr-En & GDF & En-Ro & Ro-En & GDF \\ 
\hline
\model{Fastalign} & 15.1 & 16.2 & 14.2 & 33.3 & 32.9 & 30.4\\
\model{HMM Giza++} & 11.9 & 11.9 & \textbf{8.5} & 33.3 & 36.3 & 32.4\\
\hline 
\model{HMM+NN}            & 11.8 & 11.1 & 9.7 & \textbf{30.6} & 40.1 & 34.3 \\ 
\model{HMM+NN+BPE}    & \textbf{9.8} & \textbf{10.4}  & 9.1 & 34.4 & \textbf{29.3} & \textbf{29.4} \\ \hline 
\model{HMM+VAE+BPE}   & 18.9 & 12.9  & 13.9  & 50.2 & 38.6 & 42.7 \\ 
\model{+SP}       & 12.9  & 12.2 & 11.7  & 37.5 & 38.0 & 37.0 \\ 
\model{+AC}       & 11.4  & 10.8 & 9.6  & 35.5 & 38.8 & 35.1 \\ 
\hline 
\end{tabular} 
\caption{AER scores for variants of the HMM model and for \model{Fastalign}.}
\label{tab:AER_HMM}
\end{table}

\section{Error Analysis} \label{sec:error_analysis}
\subsection{Balancing the terms in the VAE objective \label{ssec:imbalance_objective_func}}
One well-known issue of VAEs for text applications is \emph{posterior collapse} \citep{Bowman16generating,Irina17beta}\fyFuture{Add references}, where the variational distribution collapses towards the prior distribution. 

\begin{figure}[h]
\resizebox{\textwidth}{!}{
\begin{tabular}{ccc}
\includegraphics{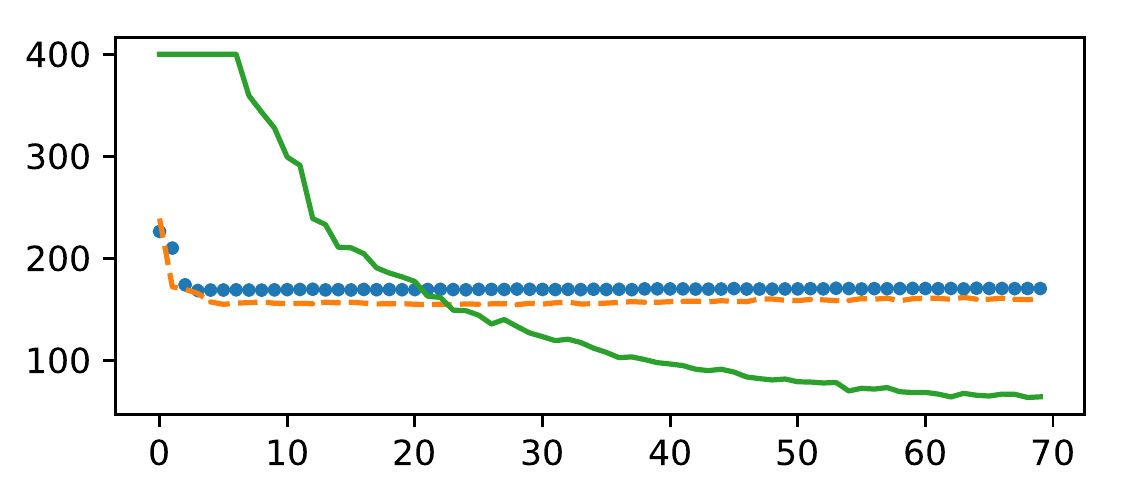} &  \includegraphics{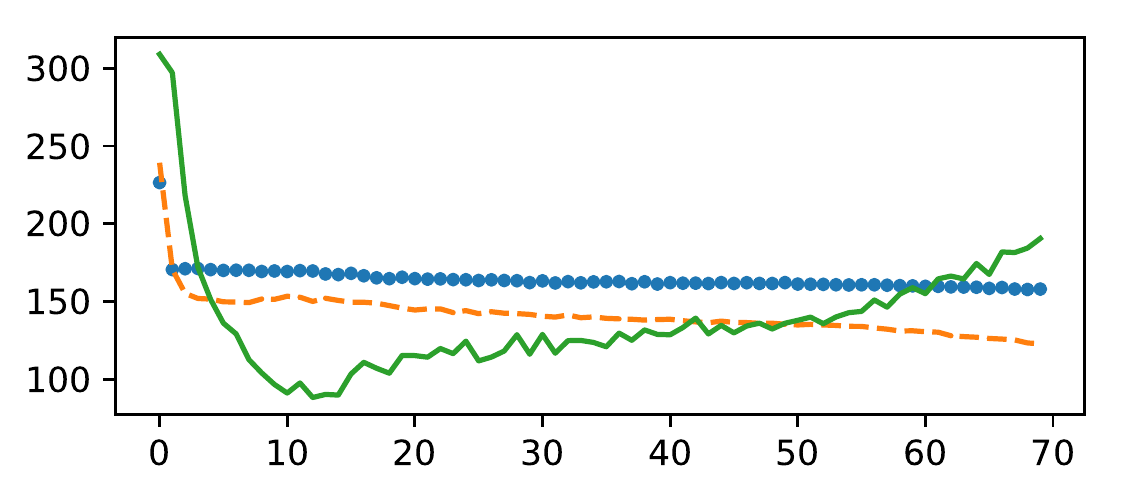} &  \includegraphics{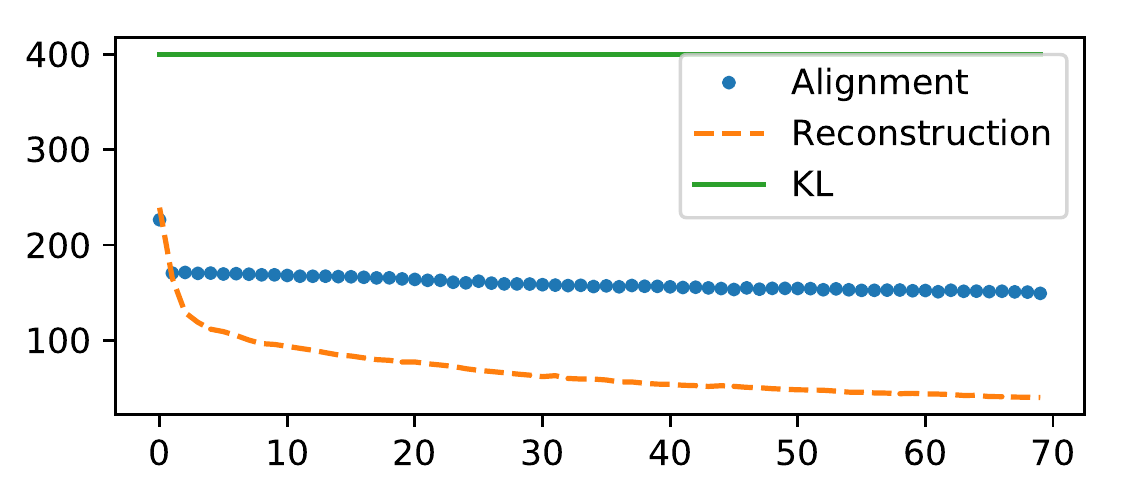} \\
 \Huge{$\alpha = 1, \beta=1, \gamma=1$} &
 \Huge{$\alpha = 1, \beta=1, \gamma=0.5$} &
 \Huge{$\alpha = 10, \beta=50, \gamma=0.5$} 
  \end{tabular}}
   \caption{Visualizing the three terms of the ELBO for Romanian-English. The weights of the reconstruction cost, alignment cost and KL divergence are set to $\alpha$, $\beta$, $\gamma$ respectively.}
 \label{fig:unbalance_cost}
 \end{figure}

This is because the KL term can get arbitrarily small, with a moderate effect on the reconstruction cost, assuming a strong reconstruction model (a recurrent network in typical applications). We also encountered this problem in our setting, but the interpretation is a bit different: when the KL term goes to zero, all words in the dictionary become indistinguishable and the reconstruction costs reaches its maximum, corresponding to the entropy of the uniform distribution of the target vocabulary. The difference in dynamics between these scores is observed in Figure~\ref{fig:unbalance_cost} (left), where we apply weights equal to $\alpha$, $\beta$ and $\gamma$ respectively to the reconstruction cost, the alignment cost and the KL divergence term. This effect is mitigated if we proportionally decrease the weight of the $KL$ term (middle). This second graph reveals the need to also better balance the importance of the other two terms. Using larger weights for the reconstruction term $(\alpha=10)$ and even more for the alignment term $(\beta=50)$, we keep the KL divergence high and make sure that the optimization focuses on decreasing the two other terms
\footnote{In our baseline experiment with English-Romanian (From http://www.statmt.org/wpt05/), using these weights resulted in an acceptable AER scores and seemed appropriate for our further experiment; a small exploration of the hyper-parameter space showed that these results were stable.}.
\akDone{}
\fyFuture{Do meta parameter search with Improve discussion}\fyDone{Attention surapprentissage}

 
\subsection{Unaligned words}
In asymmetrical models, the number of links that are generated is constant and equal to the total number of ``source'' words. A source word is deemed unaligned when it is linked to the special NULL token on the target side; a target word is unaligned when it emits no source word. We perform an in-depth analysis of these special links. Results for the alignment from French into English are in Table~\ref{tab:unaligned_word}; we observe similar trends for the other direction and for the other language pair. \fyDone{Explain accuracy}
We compute the alignment accuracy as the proportion of words (on both sides) for which the binary decision (aligned or non-aligned) is correct; we also report the precision and recall for unaligned words. Results in Table~\ref{tab:unaligned_word} show that the number of unaligned words varies in great proportion, with a minimum of about 3000 words (\model{HMM+NN}) and a maximum of nearly 6600 (\model{IBM1+VAE+BPE} and \model{HMM+VAE+BPE}). For this language pair, the reference contains 821 unaligned words. They also demonstrate the inability of all models to correctly predict null links, the best model achieving a precision of only 13.1\%.

\begin{table}[h!]
\center
\begin{tabular}{ l | c c c c c c }
Model & \# Unaligned & Accuracy & Precision & Recall \\ 
\hline 
  \model{IBM1+NN}        & 3836 & 74.0 &  8.7 & 49.7 \\
  \model{IBM1+NN+BPE}    & 3633 & 75.8 & 10.1  & 54.4 \\
  \hline 
  \model{IBM1+VAE+BPE}   & 6596 & 57.4 &  7.5 & 73.2 \\
  \model{+SP}            & 5621 & 64.2 & 9.0 & 75.1  \\
  \model{+AC}            & 5622 & 64.3 & 9.1 & 76.0  \\
  \hline 
  \hline 
  \model{HMM+NN}         & 2994 & 80.4 & 13.1  & 58.1   \\
  \model{HMM+NN+BPE}     & 4835 & 70.7 & 12.2  & 87.5    \\ 
  \model{HMM+NN+BPE+Joint}& 4843 & 70.3 & 11.6  & 83.5    \\ 
\hline 
  \model{HMM+VAE+BPE}   & 6591 & 58.6 & 8.7  & 84.9   \\
  \model{+SP}           & 6581 & 59.0 & 9.1  & 89.3   \\
  \model{+AC}           & 5579 & 65.5 & 10.4 & 86.0    \\
\hline
\end{tabular} 
\caption{Evaluation of null-alignment links when aligning French with English.}
\label{tab:unaligned_word}
\end{table}
Predicting so many unaligned words is extremely detrimental to the performance of the two basic VAE models for which we observe a very poor recall for non-null links, which is hardly compensated by the good precision scores. We see here clearly the effect of the symmetrization constraints (especially for the HMM model) where the reward associated with symmetric predictions reduces the tendency to align French words with the NULL English, and to leave too many English words unaligned. Even there (\model{HMM+VAE+BPE+SP+AC}), the number of predicted non-null links is about half as what we see for \model{HMM+NN}: as it predicts much more links than the others, this model also as an clear edge when it comes to post-hoc symmetrization, since the ``grow-diag-final'' heuristics heavily depends on the size of the intersection. Note that this problem has a much stronger overall effect in English-Romanian than in English-French. This is because the English-Romanian test set only contains sure links, which means that a low recall for aligned words directly impacts the AER. We do not see this for the French-English data, which contain many possible links that have no impact on recall \citep{Fraser07measuring}.

Incidentally, we also observe a null-word problem for \model{HMM+NN+BPE}; presumably splitting words in small units that are unrelated across languages can also make the model prefer the null alignment over links between actual words. These results clearly point out one deficiency of the current approach: for lack of having a proper model for the latent representation of the NULL token, the VAE-based approach tends to leave too many words unaligned.

\subsection{Symmetrization and agreement \label{ssec:symmetrization}}
We now study the effects of sharing parameters across alignment directions. We consider the English-Romanian test, for which the relationship between precision, recall and AER is straightforward.  Detailed scores for all \model{IBM-1} models are in Table~\ref{tab:align_word}. We see the clear benefits of sharing parameters, which contributes a jump of both precision, recall and F-measure compared with the baseline VAE. Models \model{SP} and \model{SP+AC} generate more alignment links (about +500 links) than the baseline model. This enhancement helps to outperform \model{Giza++} but is insufficient to surpass the conventional neural network models, especially when using BPE. Numbers in Table~\ref{tab:align_word} show that the gain in recall is largest in the direction En-Ro: this is because the better reconstruction of English words boosts the translation model.

\begin{table}[h!]
\center
\resizebox{\textwidth}{!}{
\begin{tabular}{ l | c c c | c c c | c c c}
Model  & \multicolumn{3}{c}{Precision} & \multicolumn{3}{c}{Recall} & \multicolumn{3}{c}{F-measure} \\
\model{IBM-1} & En-Ro & Ro-En & GDF & En-Ro & Ro-En & GDF & En-Ro & Ro-En & GDF \\
\hline
\model{Giza++}             & 58.8  & 49.9 & 73.8 & 35.1 & 43.5 & 36.5  & 43.9 & 46.4 & 48.8 \\ \hline
\model{+NN}          & 57.7  &  60.0 & 75.7 & 50.0  & 50.9 & 51.9  & 53.6  & 55.1 & 61.6 \\
\model{+NN+BPE}   & 63.9  & 64.1 &  80.4 & 50.6 & 55.6 & 55.3   & 56.5  & 59.5 & 65.5 \\ \hline
\model{+VAE+BPE}  & 56.6  & 53.9 & 79.5 & 35.4  & 37.6  & 35.0  & 43.6 & 44.3 & 48.6    \\
\model{+SP}                 & 60.6  & 57.8 & 76.2  &  43.5  & 41.8  & 42.7  & 50.7 & 48.5 & 54.8   \\
\model{+AC}                & 61.3   & 58.9 & 76.9 &  43.5 & 44.6  & 44.8  & 50.8 & 50.8 &  56.6  \\
\hline 
\end{tabular} }
\caption{Precision, recall and F-measure of IBM-1 models for English-Romanian}
\label{tab:align_word}
\end{table}
 
We now measure more directly the level of agreement between the two alignment directions for English-French (Table~\ref{tab:agreement}).
We note that the model integrating agreement costs (\model{+SP+AC}) leads to a higher number of agreements in comparison to the other VAE-based models, and also yields the best scores in terms of intersection AER.\fyDone{We need to discuss these numbers.}\fyDone{Use AER scores}

\begin{table}[h!]
\center
\begin{tabular}{ l | c c c c c c }
Model \model{HMM} & \# Agree & Ratio En-Fr & Ratio Fr-En  & AER (inter) \\ 
\hline 
\model{Giza++}       & 4683  & 72.6 & 75.5   & 7.5 \\ \hline
\model{+NN}            & 4771   & 73.2 & 76.7   & 7.4 \\
\model{+NN+BPE}    & 4040 & 75.0 & 80.2   & 10.4 \\ \hline 
\model{+VAE+BPE}   &  3160 & 69.1 & 76.0   & 18.7 \\
\model{+SP}          & 3586 & 86.2 & 86.5  & 13.0 \\ 
\model{+AC}         & 3989 & 83.6  & 84.8 & 10.1 \\ 
\hline 
\end{tabular} 
\caption{Agreement between alignments in two directions for English-French, in terms of the number of alignment links, its ratio to the total number of alignment links predicted by the model and the AER of the ``intersection'' heuristic.}
\label{tab:agreement}
\end{table}

\subsection{Training with monolingual data \label{ssec:monolingual}}
A last extension concerns the use of monolingual data in the low-resource condition. To compute the performance of the reconstruction model (R-ACC), we compute the proportion of words for which the model's prediction actually corresponds to the correct word.
Experiments are performed with English-Romanian.\footnote{The Romanian corpus is from News Crawl 2019, the English corpus is from Europarl, and corresponds to the English side of the English-French data.}
Results in Table~\ref{tab:mono_corpus} show that \model{+Mono} helps improve the reconstruction model, which attains almost perfect reconstruction accuracy in both directions, suggesting that the autoencoder is overfitting. The gain brought by monolingual data is found only for \model{IBM-1}, for the direction Ro-En (-3.6 AER). The extra-task of denoising the input (\model{+Mono+Noise}) further improves the AER compared to the parameter sharing approach. 

\begin{table}[h!]
\center
\begin{tabular}{ l  | c c | c c }
 Model & \multicolumn{2}{ c}{English-Romanian} 
 & \multicolumn{2}{|c }{Romanian-English} \\ 
\cline{2-5} 
\model{+VAE+BPE+SP} & R-ACC & AER & R-ACC & AER \\ 
\hline
\model{IBM-1}  & 84.6 & 49.3     & 93.0  & 51.4 \\
\model{+Mono} & 98.1 & 49.1 & 98.1 & 47.8 \\
\model{+Noise} & 98.4 & 48.8      & 97.9 & 47.6 \\
\hline
\model{HMM}  & 95.5  & 37.5 &  97.5  & 38.0 \\
\model{+Mono} & 98.5 & 37.9 & 98.1 & 38.0 \\
\model{+Noise} &  98.8 & 36.3 & 97.5 & 36.5 \\

\end{tabular}
\caption{Training with a monolingual corpus (\model{+Mono}) and the noise model (\model{+Noise}) on English-Romanian data. R-Acc is the accuracy of the reconstruction model.}
\label{tab:mono_corpus}
\end{table}

\fyDone{AER is F-Measure}
\fyDone{R-accuracy for the other model - there is no joint model ?}

\section{Related work\label{sec:related_work}}
The majority of recent approaches to neural word alignment fall into two categories: heuristic and probabilistic. A representative heuristic approach is \citep{Legrand16Neural}, which learns association scores between source and target word embeddings without any underlying probabilistic model. This simple approach is used to clean up translation memories in \citep{Pham18fixing}. More recently \citep{Sabet20simalign} directly takes pre-trained non-contextual and contextual multilingual representations \citep{Devlin19bert} as their association scores, deriving individual word alignments by solving an optimal matching problem.

Early work on probabilistic neural alignment is \citep{Yang2013Word}, where a feed-forward neural network is used to replace the count-based translation model of a HMM-based aligner. This approach is further developed in \citep{Tamura2014Recurrent} where a recurrent network helps to capture contextual dependencies between alignment links. This early work aims to improve the alignment quality for phrase-based MT. As discussed above, the work of \citep{Rios18deepgenerative} also considers neural versions of IBM models, with the goal to improve word representations through cross-lingual transfer in low-resource contexts. Alignment is also the main focus of \citep{Ngoho19neural} which reviews a whole set of alternative parameterizations for neural IBM-1 and HMM models, varying the word embeddings (word and character based), the context-size in the translation model and the parameterization of the distortion model.

A much more active line of research tries to improve neural MT by exploiting the conceptual similarity between alignments and attention \citep{Koehn17sixchallenges}. \citet{Cohn16alignment} modify the attention component to integrate some biases that are useful in alignements: a preference for monotonic alignements, for reduced fertility values, etc. They also propose, following \citep{Liang06alignment}, to enforce symmetrization constraints, an idea also explored in \citep{Cheng16agreement}; The same methodology is studied in~\citep{Luong15effective,Yang17neural}, with the objective to introduce dependencies between successive attention vectors. The work of \citet{Peters19sparse} also aims to enhance the attention component of a sequence-to-sequence, by enforcing sparsity via the sparse-max operator.

The work reported in \citep{Alkhouli16alignmentbased,Wang17Hybrid} explores ways to explicitely introduce alignments in NMT. They study various neuralizations of the standard generative alignment models, and also consider ways to exploit weak supervision from count-based models. This line of research is pursued by \citep{Kim17structured,Deng18latent}, where attention vectors are handled as structured latent variables in NMT; in this study, variational autoencoders are used represent the alignment structure. Finally, \citet{Garg19jointlylearning} propose to jointly learn alignment and translation in a multi-task setting, thereby improving a Transformer-based model.

When compared to heuristic approaches, an obvious defect of IBM models is their directionality, which means that they deliver asymmetric alignments. Attempts to remedy this problem, while preserving the sound probabilistic underlying models have been many. \citet{Liang06alignment} propose to jointly train EM in both directions, enforcing directional link posteriors to agree as much as possible through an additional agreement term; this work is generalized in \citep{Liu15generalized}. \citet{Graca10learning} use a different technique and enforce symmetry via additional constraints on the posterior link distribution.

Since their introduction in \citep{Bowman16generating}, VAE models of text generation have been developed in multiple ways, and applied to many NLP tasks, in particular to Machine Translation \citep{Zhang16variational}. This approach generalizes the basic VAE approach by making the latent variable and the target sentence conditionally dependent from the observed source. One major difference with our work in that the model includes one latent variable per sentence, where we consider one for each target word.
\fyFuture{More VAEs ?  \cite{Bahuleyan18variational}}

\section{Conclusion and outlook} \label{sec:perspective_conclusion}
In this paper, we have revisited the proposal of \cite{Rios18deepgenerative} and explored variants of the variational autoencoder models for the unsupervised estimation of neural word alignment models.
Our study has confirmed the previous findings and highlighted two promising aspects of this model. First, it is a full model of the joint distribution, which makes it easy and natural to introduce symmetrization constraints, as we have shown by proposing two such extensions. With these constraints, we were experimentally able to close the gap with strong baselines implementing neural variants of the conditional HMM models in the large data condition. Second, it opens new alleys to also incorporate monolingual data during training, which might especially prove useful in low-resource scenarios.

One remaining problem in this approach is the prediction of the null links, which is quite problematic in an encoder-decoder approach. We have shown in particular that the VAE model is strongly inclined to under-generate alignment links, which is detrimental to the overall AER performance. Symmetrization is a first answer to this problem, which however only partly fixes the issue. Another difficult problem with this model is controlling the optimization problem, a difficult task when the objective functions combines multiple terms with varying dynamics. More work is needed there to design better optimization strategies, with a better balance between the various sub-objectives.

\section*{Acknowledgements}
This work has been made possible thanks to the Saclay-IA computing platform.

\bibliographystyle{apalike}
\bibliography{amta2020}

\end{document}